\documentclass[10pt]{article}

\usepackage[utf8]{inputenc}
\usepackage[T1]{fontenc}
\usepackage{array}
\usepackage{hyperref}
\usepackage{graphicx} 

\usepackage[style=authoryear]{biblatex}  

\newcommand{\keywords}[1]{%
  \vspace{0.5cm}\par\noindent
  \textbf{Keywords:} #1
}

\title{Rethinking Scale: The Efficacy of Fine-Tuned Open-Source LLMs in Large-Scale Reproducible Social Science Research
}

\author{Marcello Carammia\footnote{University of Catania, Department of Political and Social Sciences, Catania, 95131, Italy}, Stefano Maria Iacus\footnote{Corresponding author: \href{siacus@iq.harvard.edu}{siacus@iq.harvard.edu}. Harvard University, Institute for Quantitative Social Science, Cambridge, MA 02138, USA}, Giuseppe Porro\footnote{University of Insubria, Department of Law, Economics and Culture, Como, 22100, Italy}}

\begin{document}

\maketitle
%
%


\begin{abstract}
 Large Language Models (LLMs) are distinguished by their architecture, which dictates their parameter size and performance capabilities. Social scientists have increasingly adopted LLMs for text classification tasks, which are difficult to scale with human coders. While very large, closed-source models often deliver superior performance, their use presents significant risks. These include lack of transparency, potential exposure of sensitive data, challenges to replicability, and dependence on proprietary systems. Additionally, their high costs make them impractical for large-scale research projects.

In contrast, open-source models, although available in various sizes, may underperform compared to commercial alternatives if used without further fine-tuning. However, open-source models offer distinct advantages: they can be run locally (ensuring data privacy), fine-tuned for specific tasks, shared within the research community, and integrated into reproducible workflows.

This study demonstrates that small, fine-tuned open-source LLMs can achieve equal or superior performance to models such as ChatGPT-4. We further explore the relationship between training set size and fine-tuning efficacy in open-source models. Finally, we propose a hybrid workflow that leverages the strengths of both open and closed models, offering a balanced approach to performance, transparency, and reproducibility.

\end{abstract}

\keywords{large language models, fine-tuning, text classification, reproducibility, computational social science}

\section*{Introduction}
The widespread availability of generative AI, particularly large language models (LLMs), is transforming both society and science. These models are becoming increasingly easy to prompt, making them appealing to scientists across both the hard and social sciences for conducting text classification tasks in a wide range of disciplines, including  finance \parencite{Loukas23}, health studies \parencite{Guo24,cao2024,Kiyasseh24}, environmental studies \parencite{trajanov2023},
geoscience \parencite{maze2024},
physics \parencite{li2024},
chemistry \parencite{liao2024}
radiology \parencite{Nowak2024}, psychology \parencite{Suhaib24}, sociology \parencite{kozlowski2024,Mutzel23,Chae2023}, political science \parencite{LiuGe2024}, criminology \parencite{Nikolakopoulos24}, law studies \parencite{Wei23}, just to mention a few.  Recent studies have sparked both enthusiastic \parencite{Gilardi2023,rouzegar2024} and critical \parencite{Suhaib24} assessments of LLMs.

A central debate in this field involves closed vs. open foundation models \parencite{Zhang24,Nowak2024,Suhaib24,hanke2024open,Chae2023}. Proprietary models, such as those from OpenAI and Google, present several challenges: they can change versions without notice, their architectures are not fully disclosed, and they may be subject to content moderation—such as filtering for hate speech or privacy concerns—designed to protect companies from legal and reputational risks. While these safeguards are important, they fall outside the control of scientists.

For researchers, the financial and computational resources required to train large proprietary models, like OpenAI's GPT-4 (estimated to cost \$100M to develop), are prohibitive. However, smaller, open-source models can be fine-tuned on readily available hardware, such as a laptop or a small GPU cluster, making them feasible for academic projects. Even companies like OpenAI are now moving towards smaller, more efficient models using techniques like the mixture-of-experts approach.

LLMs are typically measured by the number of parameters in their neural network architecture. While proprietary models often withhold exact specifications (e.g., OpenAI’s GPT-4 is estimated to use over 1.7 trillion parameters, Google's Gemini Ultra has probably 1560B parameters and Anthropic's Claude 3 is rumored to have 500B parameters), open-source models provide clear documentation of their architectures. This study focuses on the Meta-developed LLAMA family of open-source models, including versions from 7B to 405B parameters. We do not consider other open-source models, such as Mistral/Mixtral, BLOOM, and Falcon, although they share similar fine-tuning capabilities and usage restrictions based on licensing.

Fine-tuning allows these open-source models to be specialized for specific tasks without retraining from scratch. In this study, we fine-tuned the LLAMA models using Low-Rank Adaptation \parencite{hu2021} (LoRA) , which modifies the model's output to better suit particular domains or tasks. Unlike proprietary models such as ChatGPT, whose fine-tuned versions remain under corporate control, open-source fine-tuned models can be freely shared, advancing reproducible (social) science \parencite{TrustMeBro}.

This research was motivated by the need to analyze Harvard's 10B Geo-Tweets archive \parencite{Lewis16}. At current costs, applying commercial LLMs to this scale would be unfeasible, with token pricing reaching hundreds of thousands of dollars for large-scale analysis. In contrast, open-source models can run on local clusters, using tools like LLAMA.cpp \parencite{llamacpp},  Oolama \parencite{ollama},  GPT4All \parencite{gpt4all} or similar open source projects, making large-scale, affordable analysis possible.

In this work, we demonstrate that while larger models generally outperform smaller ones, fine-tuning can level the playing field. Properly fine-tuned open-source models can achieve performance comparable to, or better than, significantly larger models like ChatGPT-4, especially in non-trivial classification tasks. Additionally, we examine how the size of the training data impacts the effectiveness of fine-tuning, finding that models trained on more extensive data are harder to fine-tune effectively.


Finally, we introduce a hybrid approach that utilizes both proprietary models, like ChatGPT-4, to accelerate the creation of labeled datasets for fine-tuning, in cases where labeled data is not readily available.

We evaluate these approaches through three classification tasks: (i) classifying tweets into 46 dimensions of well-being from the Human Flourishing Program \parencite{VanderWeele2017}; (ii) classifying European Parliamentary questions into 19 policy areas from the Comparative Agenda Project \parencite{Alexandrova14,CAP} (CAP); and (iii) classifying datasets from the Harvard Dataverse repository into 15 subject categories. These tasks, ranging from noisy, unlabelled data to highly curated datasets, extend beyond simple sentiment analysis and showcase the capabilities of fine-tuned open-source models.

\bigskip
\section*{Related Work}
Numerous studies \parencite{Chae2023,pangakis2024,rouzegar2024,yin2024,Gougherty2024} have shown that recent advancements in large language models (LLMs) have revolutionized so much the field of natural language processing that they now  hold a significant potential for social science research. Since the introduction of pioneer LLMs like BERT \parencite{devlin-etal-2019-bert} (a model with ``just'' 110 million parameters) up to XLM-RoBERTa \parencite{conneau-etal-2020-unsupervised} (550 million parameters), the development of LLMs has progressed in the last few years through two key advancements: refined Transformer architectures \parencite{Vaswani17} and optimized training techniques.
These improvements, along with increased parameter sizes in models like GPT-3 and T5, have enabled LLMs to capture complex language patterns across diverse contexts, making them viable for varied social science tasks with minimal fine-tuning \parencite{brown2020,Raffel2020}.

With LLMs increasingly accessible as research tools, computational social science \parencite{lazer2009} has started to leverage these models for large-scale text analysis beyond basic sentiment classification. This shift in scale has spurred numerous studies examining LLMs for annotation, classification, and knowledge extraction from text \parencite{pangakis2024,Singh2024,zhang_etal_2024,fechner2024,rouzegar2024,Bisbee2024,Barrie2024}. Recently, researchers have also explored the multi-modal capabilities of LLMs for combining text with other data types \parencite{Miah2024}. Our work aligns with studies examining fine-tuning’s impact on LLM performance, where findings consistently indicate that fine-tuned models can handle specific classification tasks with high accuracy \parencite{yin2024,mathav2024,bucher2024}.
 What these works have in common is that they consider simple classification tasks (like sentiment or very few categories) and a limited set of LMMs, but they come to conclusions in agreement with our analysis. 
  
The debate over proprietary versus open-source models has grown as researchers have increasingly voiced concerns about access, transparency, and scalability \parencite{Nowak2024,Nowak24-german,Wang2024,Zhang24,hanke2024open,Fields24,Palmer24}. This is indeed becoming  a real issue in science as there is yet no way for individual researchers, or even a consortium of academical institutions, to replicate the scale of the training that commercial players are able to perform. 
    We think the best of the two worlds can be leveraged to do better, faster and more scalable social science research.
 While proprietary models offer advanced generalization capabilities due to extensive training, open-source LLMs, despite smaller in size, are accessible for fine-tuning on specific tasks. 
 Scholars increasingly recognize that combining task-specific tuning with open-source models can achieve performance competitive with proprietary models \parencite{irugal2024,xiang_li_2023,fu2024}. 
It is also a shared and emerging opinion among scholars and methodologists in the social science that only open-source models can guarantee a level of what is usually referred to as \textit{replicability} (or \textit{reproducibility}), i.e., ``deterministic code and
 data" replication \parencite{TrustMeBro}. Conversely to these authors though in this work we show that these open source models are not just ``last resort'' tools as they can achieve equal or better accuracy than black-box models if effectively fine-tuned.

 Such approach enable cost-effective, reproducible research, bridging the gap between proprietary and open-source capabilities by leveraging field-specific knowledge and specialized datasets. Our research underscores that integrating both proprietary and open-source models can optimize even more performance, scalability, and accessibility in social science research.

\section*{Methods}
\subsection*{Streamlining Labeled Dataset Creation for Fine-Tuning and Scaling}
The traditional approach to text classification involves training human coders, assigning them random sets of documents to classify, and resolving disagreements among coders to generate a clean set of labeled data for training and validation. 
An empirical fact familiar to researchers is that human coders often disagree in 15-20\% of cases. This disagreement arises from various factors, such as inflexible codebooks, cultural and political biases, coder fatigue, or the inherent uncertainty in capturing certain dimensions.

Achieving high inter-coder agreement can be challenging, and surprisingly, coders often prefer LLM-generated annotations over expert answers in certain contexts \parencite{Goyal2023,Hagendorff2023}. In some studies, the correlation between annotations by platforms like Mechanical Turk and expert annotators can approach zero \parencite{fabbri2021}. Recent literature has also questioned the reliability of human classification, traditionally considered the ``gold standard'', suggesting that it may not be as accurate as commonly assumed \parencite{Clark2021,Liu23}.

Another critical issue is scalability. Creating a sufficiently large set of classifications to cover both training and validation datasets is time-consuming and expensive. In cases like ours, which involve a high dimensionality of categories, even expert coders find the task challenging. Furthermore, using human cognitive resources for this task may not be the most efficient use of time and skill.

Our proposed approach is to generate multiple classifications for an unlabeled set using potentially several LLMs, with humans focusing on rejecting incorrect classifications. It's faster and easier to spot what does not apply to a text—especially when faced with a set of over 10 possible labels (our application uses 15, 20, and 46 labels)—than to identify which set of labels apply.

There are several psychological reasons why it’s often easier for humans to recognize errors rather than identify correct classifications. For example, \textit{error salience} \parencite{Harsay2012} suggests that mistakes or anomalies tend to stand out more than correctly processed information, which blends into our expectations. \textit{Human cognitive biases}, like being more sensitive to inconsistencies (an ancestral survival mechanism that makes us more sensitive to risks or issues), also play a role \parencite{Mobbs2015}. Lastly, the \textit{negativity bias} \parencite{ito1998negative} makes humans pay more attention to negative or incorrect information than to positive or correct information.

All the above motivates our approach on how to accelerate the creation of a labeled set.

\begin{figure}[ht]
\centering
\includegraphics[width=0.9\linewidth]{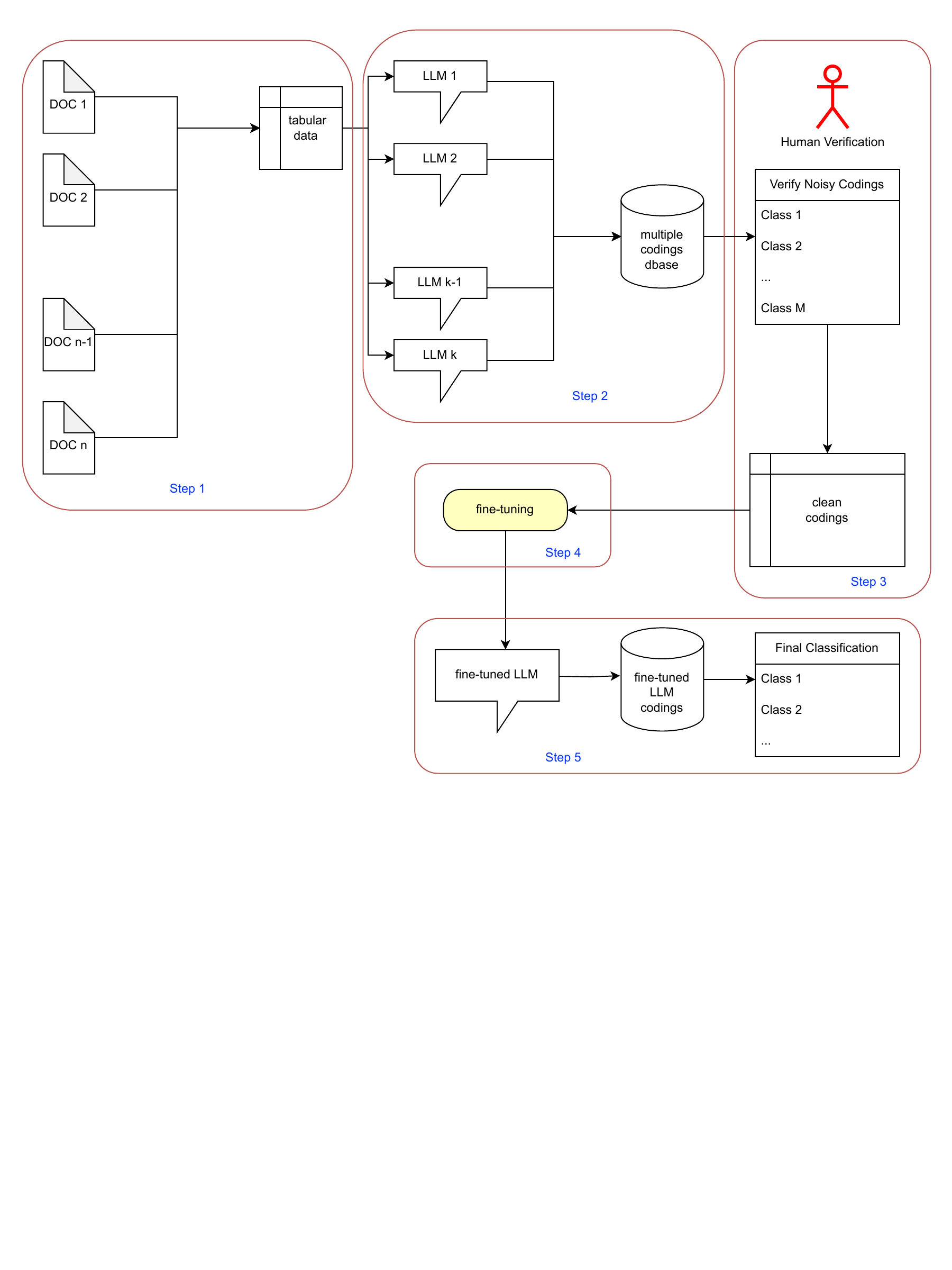}
\caption{Acceleration of labeled datasets creation for fine-tuning using multiple LLMs and human verification.}
\label{fig:workflow}
\end{figure}

The proposed workflow is presented in Figure~\ref{fig:workflow} and is composed of five steps. We assume the general context of multi-label classification.

\begin{itemize}
    \item \textbf{Step 1: \textit{Data Normalization}}. This step is common to all classification tasks and involves organizing the documents in a corpus into a tabular format (such as a CSV file or an SQL table in a database). The data contains only an identifier, \texttt{id}, and a \texttt{text} field. If the actual label is available, a \texttt{true} label field is also included.
    
    \item \textbf{Step 2: \textit{AI-Crowd Classification}}. This step involves using one or more LLMs (or the same LLM with different prompts) to classify each document. At the end of this process, each document is labeled multiple times, and any duplicate labels are removed.
    
    \item \textbf{Step 3: \textit{Human Approval}}. Multiple coders (experts, or a combination of both) are asked to reject labels that do not apply to each document. Multiple coders may be randomly assigned to the same set of documents to assess intra-coder reliability.
    
    \item \textbf{Step 4: \textit{Fine-tuning}}. A single LLM (potentially different from the ones used in Step 2) is fine-tuned using the human-filtered labeled set.
    
    \item \textbf{Step 5: \textit{Scaling}}. The fine-tuned model is used to scale the analysis to previously unseen documents.
\end{itemize}

Each step can be expanded and tailored as needed, but these details do not alter the fundamental essence of the workflow. For example, the training set of documents can be split into training, validation, and test sets to assess the performance of different LLMs within the proposed workflow. 

After introducing several metrics to evaluate the performance of the different LLMs, we will discuss how this workflow is applied to the case studies below.

\subsection*{Evaluation Metrics}
In this work, we employ various accuracy metrics depending on the specific classification problem being addressed. Let 
$n$
 represent the total number of documents to classify, and let 
$M$
 denote the number of possible labels in a given classification task.

We define \( y_{ij} \) as an indicator function that takes the value \( 1 \) if document \( i \), where \( i = 1, \ldots, n \), is assigned to class \( j \), where \( j = 1, \ldots, M \), and \( 0 \) if label \( j \) does not apply to document \( i \).

For mutually exclusive categories, each document \( i \) can belong to at most one class, meaning there exists at most one index \( j \) such that \( y_{ij} = 1 \). In such cases, we use the notation \( y_i = j \) to indicate that document \( i \) belongs to category \( j \).

We now review a few basic metrics commonly used to assess the quality of a classifier in the context of mutually exclusive categories.

\subsubsection*{Accuracy}
If \( y_i \) is the actual true category for document \( i \) and \( \hat{y}_i \) is the category predicted by the classifier, then accuracy is defined as 
$$
\textrm{Accuracy} = \frac{1}{n} \sum_{i=1}^n \mathbf{1}(\hat{y}_i = y_i),
$$
where the function \( \mathbf{1}(\texttt{cond}) = 1 \) if the condition `\texttt{cond}' is true and \( 0 \) otherwise.

The accuracy is simply the sum of the diagonal elements of the \( M \times M \) confusion matrix \( C = (c_{jk}, j, k = 1, \ldots, M) \), where \( c_{jk} \) counts the number of times a document with true label \( j \) is assigned to category \( k \) by the classifier.
It is also worth to introduce other measures like Precision, Recall and Specificity. The Precision measure is defined as:
$$
\textrm{Precision}(\textrm{class} = j) =
\frac{\textrm{TP}(\textrm{class} = j)}{\textrm{TP}(\textrm{class} = j) + \textrm{FP}(\textrm{class} = j)} = \frac{\sum_i \mathbf{1}((\hat{y}_i =j) \cap ( y_i = j))}{\sum_i \mathbf{1}(\hat{y}_i =j)  },
$$
where TP refers to `true positives' and FP refers to `false positives.'
Similarly, the Recall (or Sensitivity) for class \( j \) is defined as:
$$
\textrm{Recall}(\textrm{class} = j) = \textrm{Sensitivity} (\textrm{class} = j) =
\frac{\textrm{TP}(\textrm{class} = j)}{\textrm{TP}(\textrm{class} = j) + \textrm{FN}(\textrm{class} = j)} = \frac{\sum_i \mathbf{1}((\hat{y}_i =j) \cap ( y_i = j))}{\sum_i \mathbf{1}(y_i = j) },
$$
where FN refers to `false negatives.' 
Finally, the Specificity measures the proportion of actual negatives that are correctly identified by the classifier. For class \( j \), Specificity is defined as:
$$
\textrm{Specificity}(\textrm{class} = j) = 
\frac{\textrm{TN}(\textrm{class} = j)}{\textrm{TN}(\textrm{class} = j) + \textrm{FP}(\textrm{class} = j)} =
\frac{\sum_i \mathbf{1}(\hat{y}_i \neq j \cap y_i \neq j)}{\sum_i \mathbf{1}(y_i \neq j)}.
$$
Another important metric is the so-called
Balanced accuracy, that is used  when  dealing with imbalanced datasets (like in our examples below). It aims to provide a better measure of performance when the classes in a classification problem are not equally represented.
The main purpose of balanced accuracy is to address the limitations of standard accuracy in scenarios where the dataset has an unequal distribution of classes. It does this by taking into account both the Sensitivity (true positive rate) and the Specificity (true negative rate), giving equal weight to both.
It is given by the formula: 
$$\textrm{Balanced Accuracy} = \frac{\textrm{Sensitivity}+\textrm{Specificity}}{2}.
$$
The Macro Balance Accuracy is the simple average of the Balanced Accuracy by class.
\subsubsection*{F1 Score}
The F1 score, originally introduced for a $2 \times 2$ table, is given by the formula:
$$
\textrm{F1} = 2 \times \frac{\textrm{Precision} \times \textrm{Recall}}{\textrm{Precision} + \textrm{Recall}}.
$$

The F1 score for class \( j \) is then computed as:
$$
\textrm{F1}_j = 2 \times \frac{\textrm{Precision}(\textrm{class} = j) \times \textrm{Recall}(\textrm{class} = j)}{\textrm{Precision}(\textrm{class} = j) + \textrm{Recall}(\textrm{class} = j)}.
$$

As with accuracy, higher values of the F1 score indicate better classifier performance.

\subsubsection*{Hamming Loss}
In multi-label classification, each document can be assigned multiple labels. In general, each document \( i \) is associated with a vector \( y_{i*} = (y_{i1}, y_{i2}, \ldots, y_{iM}) \), which indicates the presence or absence of labels. Let \( \hat{y}_{i*} \) be the predicted vector of \( y_{i*} \) for document \( i \).
The Hamming loss is a measure of how frequently a prediction is wrong. It is particularly useful in this context of multi-label classification. For a single document \( i \), the Hamming Loss \parencite{Hamming50} is given by the formula:
$$
\textrm{HL}_i = \frac{1}{M} \sum_{j=1}^{M} \mathbf{1}(\hat{y}_{ij} \neq y_{ij}).
$$
The overall Hamming Loss for the sample of documents is:
$$
\textrm{HL} = \frac{1}{n} \sum_{i=1}^n \textrm{HL}_i = \frac{1}{nM} \sum_{i=1}^n \sum_{j=1}^{M} \mathbf{1}(\hat{y}_{ij} \neq y_{ij}).
$$
The metric \( \textrm{HL}_i \) is equal to 1 if all predicted \( y_{ij} \) values are wrong, and 0 if all predicted \( y_{ij} \) values are correct. The overall Hamming Loss, which is just the average of the \( \textrm{HL}_i \)'s, also ranges in \([0, 1]\). When comparing two classifiers, the one with lower Hamming Loss is better.

\subsubsection*{Jaccard Index}
Another metric that is popular in multi-label classification problems is the Jaccard Index \parencite{Gilbert1884,Jaccard1901}. It measures the similarity between the predicted and true vectors of \( y_{i*} \) by dividing the size of their intersection by the size of their union. The formula is as follows:
$$
J = \frac{1}{n} \sum_{i=1}^n \frac{|\hat{y}_{i*} \cap y_{i*}|}{|\hat{y}_{i*} \cup y_{i*}|} =
\frac{1}{n} \sum_{i=1}^n \frac{|\hat{y}_{i*} \cap y_{i*}|}{M} =
\frac{1}{nM} \sum_{i=1}^n |\hat{y}_{i*} \cap y_{i*}|,
$$
where \( |A| \) is the number of elements in set \( A \).
In the above formula, the vector \( y_{i*} \) (and similarly \( \hat{y}_{i*} \)) should be treated as a set formed by the union of its vector elements, i.e., \( \{y_{i1}\} \cup \{y_{i2}\} \cup \dots \cup \{y_{iM}\} \).

The ratio between the numerator and denominator is 0 if the two vectors \( \hat{y}_{i*} \) and \( y_{i*} \) are completely different, and 1 if the vectors are identical. Thus, for the Jaccard Index, when comparing two classifiers, the one with the higher value is better. The Jaccard Index essentially plays the role of accuracy in a multi-label classification setting. While Hamming Loss focuses on individual label mismatches, the Jaccard Index emphasizes overall set similarity.

\section*{Data and Problem Statements}
We will consider three different datasets and classification problems which present a spectrum of challenges.

\begin{table}[ht]
\centering
\small
\begin{tabular}{|m{3.7cm}|m{1.3cm}|m{0.9cm}|m{0.7cm}|m{4cm}|m{4cm}|}
\hline
 Project & Verification  &Training   & Test  &  Challenges & Features\\
\hline
Human Flourishing Program&\centering 10,000 &\centering 2,404 &\centering 2,177 &  multi-label classification; high-dimensional (46 categories); severely imbalanced; noise in the data; no gold standard & very large scale application (10B tweets archive) \\
\hline
Comparative Agenda Project &\centering 1,964&\centering 1,272 &\centering 424 & mutually exclusive categories; medium/high-dimensional (20 categories); severely imbalanced & absence of noise; large scale application; gold standard exists\\
\hline
Harvard Dataverse & \centering -- & \centering 5,000 or 76,110 & 32,619 & multi-label classification; medium-dimensional  (15 categories)& highly curated; very low noise; very large training set \\
\hline
\end{tabular}
\caption{\label{tab:challanges}Challenges and features of the different projects used in this work. Verification is the set of unsupervised labeled data done by the mixture of LLMs in the workflow of Figure~\ref{fig:workflow}.}
\end{table}

\subsection*{Classification of Tweets according to the Human Flourishing Program}
The Human Flourishing Program \parencite{VanderWeele2017} is a research initiative based at the Institute for Quantitative Social Science (IQSS) at Harvard University. Its goal is to study and promote human flourishing across a broad spectrum of life domains, integrating interdisciplinary research in the social sciences, philosophy, psychology, and other fields. The program aims to understand what constitutes human well-being or \textit{flourishing}, which goes beyond mere happiness or economic success. It seeks to identify and analyze the factors that contribute to a flourishing life, including physical and mental health, happiness and life satisfaction, meaning and purpose, character and virtue, and close social relationships. Within this program, the  Global Flourishing Study (GFS) is a five-year longitudinal data collection on approximately 200,000 participants from 20+ geographically and culturally diverse countries and territories, including Argentina, Australia, Brazil, China (Hong Kong), Egypt, Germany, India, Indonesia, Israel, Japan, Kenya, Mexico, Nigeria, the Philippines, Poland, South Africa, Spain, Sweden, Tanzania, Turkey, United Kingdom, and the United States. GFS measures global human flourishing in six areas: \textit{i)} Happiness and life satisfaction; \textit{ii)} Mental and physical health; \textit{iii)} Meaning and purpose; \textit{iv)} Character and virtue; \textit{v)} Close social relationships and \textit{vi)} Material and financial stability. 

Each of these human flourishing areas is investigated through several questions. For our study, we have selected 46 dimensions (see \textbf{Supplementary Material}) among all the dimensions identified by the GFS study, taking into account the fact that we cannot ask question but we should infer the presence or absence of each dimension in a particular tweet. The scope of our study, that goes much beyond the scope of the present work, it to analyze the entire Harvard's 10B Geo-Tweets archive \parencite{Lewis16} with an accurate and scalable solution.
We will analyze the performance of open source models from the Meta-developed LLAMA family, testing base and fine-tuned versions of LLAMA-2, 3, 3.1 and 3.2 versions for different sizes of model parameters.

For this analysis we selected 10,000 tweets at random in English language. We then applied the workflow illustrated in Figure~\ref{fig:workflow}, letting four models (LLAMA-2 7B, 13B and 70B and ChatGPT4) produce the unsupervised codings of Step 2 of the workflow. Then, we trained a group master and doctoral students to verify (Step 3) the AI-classifications according to the codebook of the project (see \textbf{Supplementary Material}). From the cleaned codings we proceeded to fine-tune several of the above models (those up to 70B parameters). The cleaned labelled data at the end of the workflow, amount to a total of 4,581 tweets. The rest of 10,000 AI-classified data are left for future use and validation as human coders were only able to verify a portion of them.
We split the cleaned data into a training set and a test set. We then validate each model (both base and fine-tuned ones).

\subsection*{Policy Attribution of European Parliamentary Questions}
The Comparative Agendas Project (CAP) is an international research initiative aimed at systematically tracking, coding, and comparing policy attention and issue agendas over time across various countries \parencite{Alexandrova14,CAP}. The project seeks to understand how government attention and public priorities shift across policy domains, such as healthcare, defense, education, and the environment.

The CAP accomplishes this by analyzing policy outputs—such as laws, government speeches, media coverage, and legislative bills—coding them into standardized categories, and comparing them across time and countries. This approach enables researchers to examine how governments respond to public demands and to assess whether certain issues are prioritized over others across different contexts.

While the exact number of agenda items or categories may vary depending on updates or extensions for specific countries, the core CAP codebook typically includes 19 major policy areas, subdivided into over 200 subcategories. These major areas include: Macroeconomics; Civil Rights, Minority Issues, and Civil Liberties; Health; Agriculture; Labor, Employment, and Immigration; Education; Environment; Energy; Transportation; Law, Crime, and Family Issues; Social Welfare; Community Development and Housing Issues; Banking, Finance, and Domestic Commerce; Defense; Space, Science, Technology, and Communications; Foreign Trade; International Affairs and Foreign Aid; Government Operations; and Public Lands, Water Management, and Territorial Issues.

In this study, we focus on the classification of European Parliament questions from 1994 to 2021. The dataset comprises 174,161 questions posed over seven legislative terms, covering 28 countries and eight European party families. These questions reflect various EU institutional configurations, ranging from Maastricht to Lisbon, and are all provided in English.

According to the CAP framework, these questions can be classified into 19 broad policy areas. Certain categories, such as Defense, Foreign Trade, and Social Policy, are under-represented, while others, such as Macroeconomics, Agriculture, Civil Rights, and International Affairs and Foreign Aid, are over-represented. This imbalance poses challenges for classification. Nonetheless, the dataset is free from noise, as each parliamentary question is assigned to a specific policy area based on the CAP project’s codebook guidelines.

Although the CAP coding protocol assumes that each text should be classified into one of the 19 major policy areas, this task presents significant challenges for human coders. It is well-known among CAP project experts that inter-coder reliability typically around 70\% and sate-of-the art models \parencite{SebHok2024} built on top of very large training-set data (1,147,783 observations) show weighted F1 score that varies from 0.71 to 0.9 according to different data sources (Media,	Social Media, Parliamentary Speech,	Legislative	Executive Speech,	Executive Order,	Party Manifesto,	Judiciary), signaling that the CAP data are not easy to classify. This variability highlights the inherent difficulty of consistently assigning policy areas, even with detailed coding guidelines. On this particular dataset (parliamentary questions) the inter coder reliability over size well trained coders on a subset of 134 manually coded parliamentary questions, the Cohen's kappa statistics \parencite{fleiss1971mns} is equal to 66.4

In order to classify a preliminary set of data according to Step 2 of the workflow in Figure~\ref{fig:workflow}, we used three different strategies:
\begin{itemize}
    \item \textit{direct} method: we prompted ChatGPT4 to select one of the 200 micro areas and then aggregated by macro areas;
    \item \textit{zero shot} method: we asked ChatGPT4 to choose one of the 19 categories;
    \item \textit{iterative} method:
    \begin{itemize}
        \item for each of the 19 policy macro areas we prompted ChatGPT4 to choose only one among the  corresponding subtopic areas  or \texttt{None};
        \item after iterating the 19 policy macro areas, we are left with a subset of subtopics corresponding each to a different macro area;
        \item we map back the subtopic to their macro areas and ask ChatGPT4 to choose only one of these. In this case \texttt{None} is not an option, ChatGPT4 is forced to choose one policy area. 
    \end{itemize}
\end{itemize}
The iterated method requires 20 calls to OpenAI's API's, so it is the more expensive of the three although it's accuracy was the highest.

\subsection*{Classification of Harvard Dataverse datasets}
Harvard Dataverse (\url{https://dataverse.harvard.edu}) is a generalist research data repository \parencite{King07} with highly curated metadata. Thousands of scholars across many disciplines have deposited data, along with their associated metadata, in this repository. To support researchers in depositing their data more easily, the Dataverse development team 
is experimenting to generate metadata suggestions based on data characteristics, with the use of large language models (LLMs). In this experiment, we explored the ability of a small LLAMA-7B fine-tuned model to predict one or more subject categories for datasets using only the dataset \texttt{Title} and \texttt{Description}. The outcome of this project will be  also used to verify the correctness of metadata in already published datasets or to test their FAIR-ness \parencite{sdata201618}. Previous works \parencite{TurboCurator,shigapov2024} tried to do this using ChatGPT but again the approach is not scalable and not fully under the control of the researcher.

The dataset collection contains 13 subject categories, inherited from the Revised Field of Science and Technology (FOS) Classification by OECD \parencite{OECD_FOS_Classification_2007}, plus ``Other'' and ``N/A'' when information is missing, for a total of 15 labels to predict in this classification task. Since each dataset can belong to one or more subject categories, this is a multi-class classification problem. Notice that in this case a human supervised training set already exists, so there is no need to apply the workflow in Figure~\ref{fig:workflow}. The LLM is asked to return up to 3 subject categories per dataset. This choice is motivated by the application at hand in which the Dataverse system suggests the three most relevant subject categories that apply to the users that is submitting data to the archive without overwhelming them with too much choices. In this system the user is supposed to reject the wrong subject categories. 

For simplicity, we report accuracy based on the number of times the LLM correctly predicts at least one subject category. The subject categories are: 
``Agricultural Sciences'', 
``Arts and Humanities'',
``Astronomy and Astrophysics'',
``Business and Management'',
``Chemistry'',
``Computer and Information Science'',
``Earth and Environmental Sciences'',  
``Engineering'',
``Law'',
``Mathematical Sciences'',
``Medicine, Health and Life Sciences'',
``Physics'',
``Social Sciences'', plus ``Other'' and ``N/A''.

The Harvard archive of datasets comprises 108,729 datasets, of which 76,110 (70\%) were used as the training set, and 32,619 as the test set. These sets were selected randomly. We generated two fine-tuned versions of the LLAMA2-7B model, using respectively the entire training set and a random subset of 5,000 datasets from the larger training set. We denote with ``large'' and ``small'' the two fine-tuned model in the results. We then compared the classification performance of the two fine-tuned LLAMA2-7B models against the base versions of LLAMA2 at sizes 7B, 13B, and 70B.

\section*{Results}
\subsection*{Tweets Classification according to the Human Flourishing Program}
\begin{figure}[ht]
\centering
\includegraphics[width=\linewidth]{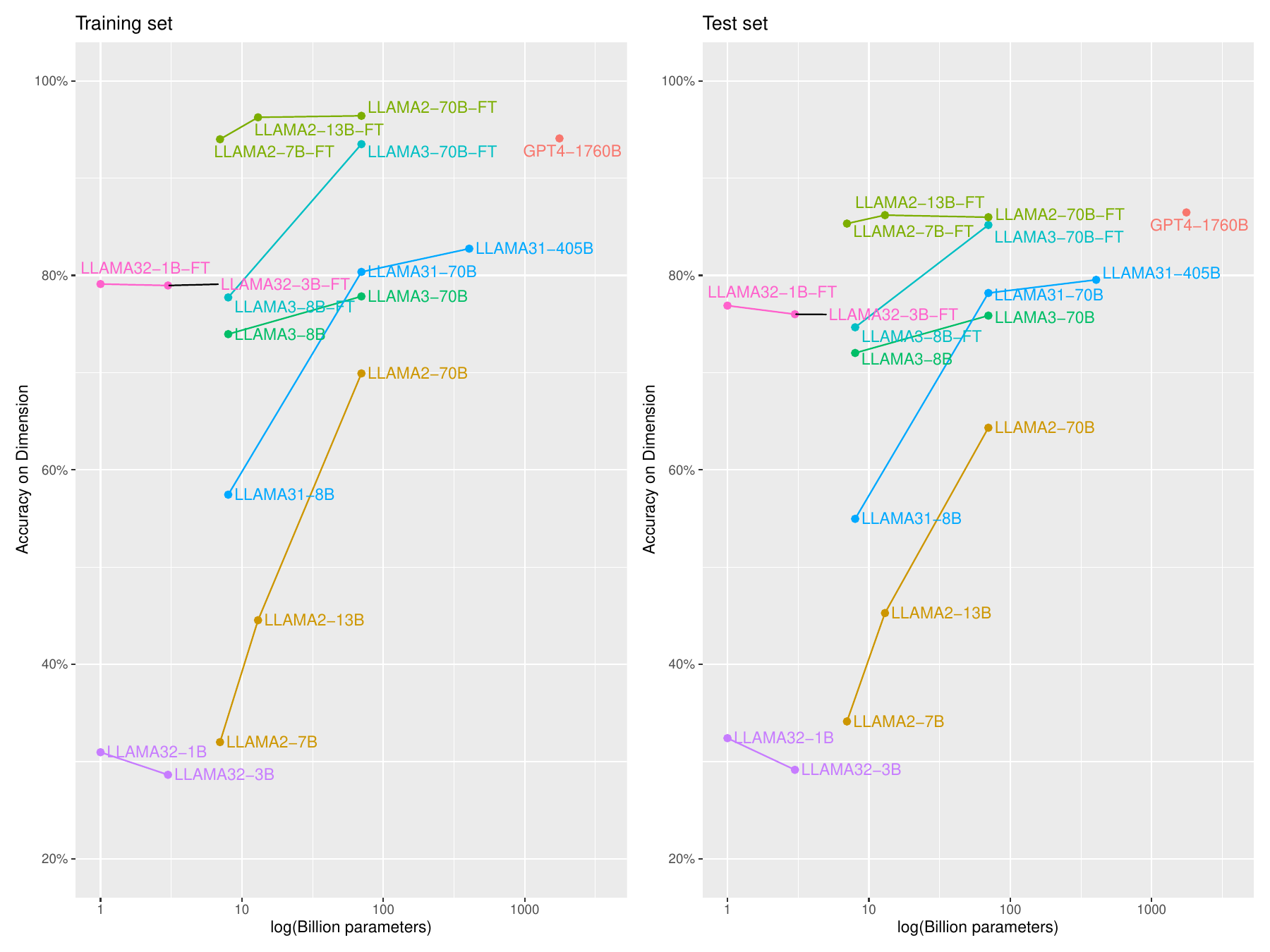}
\caption{Accuracy for different LLAMA models and ChatGPT4 on both the training and the test sets. The `FT' suffix stands for `fine-tuned version' of the same model. Number of model parameters in the log-scale. For ChatGPT the real dimension of the model is not known, but seems to be built on top of 8 models of about 220B parameters.}
\label{fig:performance1}
\end{figure}

\begin{figure}[ht]
\centering
\includegraphics[width=\linewidth]{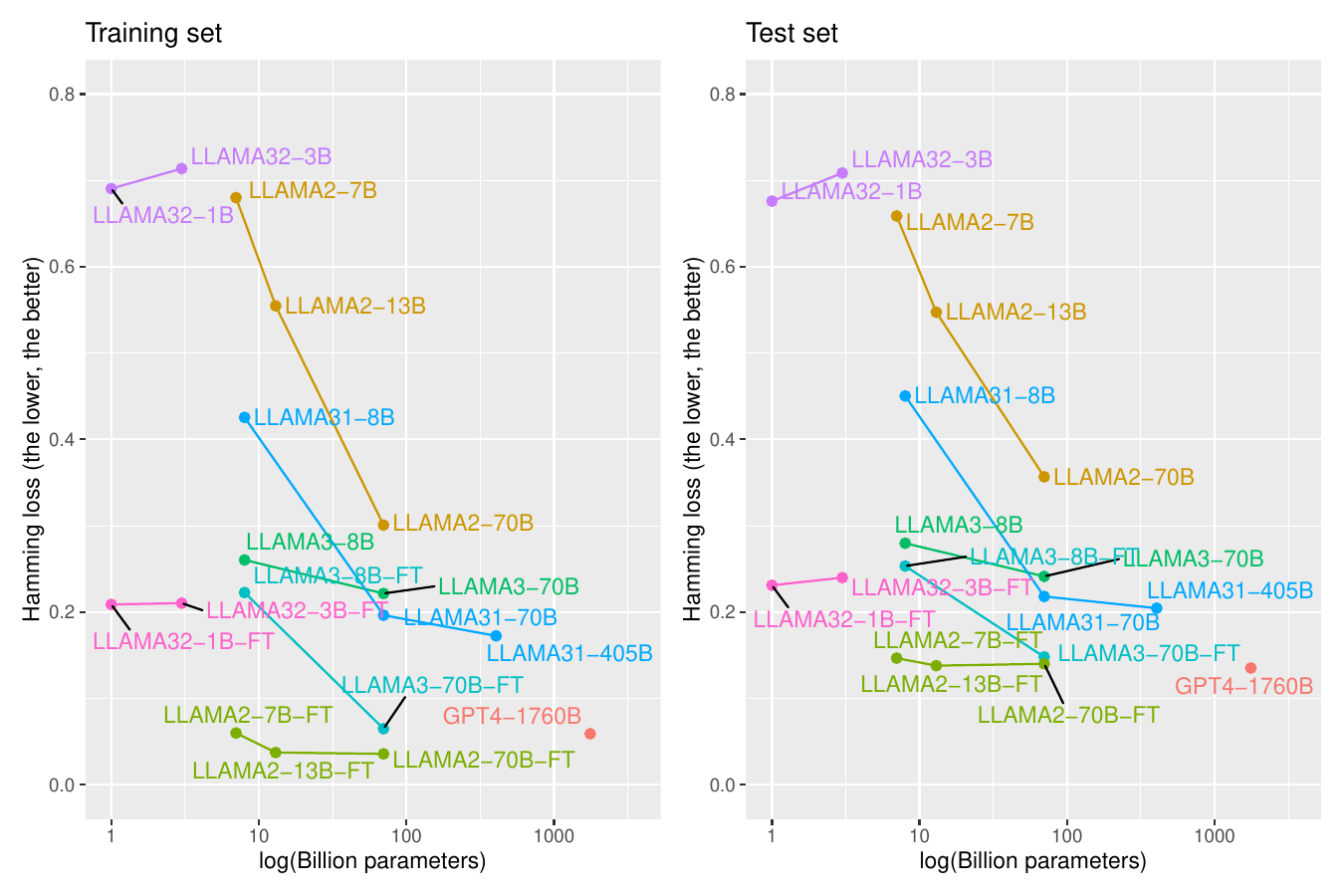}
\caption{Hamming loss for different LLAMA models and ChatGPT4 on both the training and the test sets. The `FT' suffix stands for `fine-tuned version' of the same model. Number of model parameters in the log-scale. For ChatGPT the real dimension of the model is not known, but seems to be built on top of 8 models of about 220B parameters.}
\label{fig:performance2}
\end{figure}

\begin{figure}[ht]
\centering
\includegraphics[width=\linewidth]{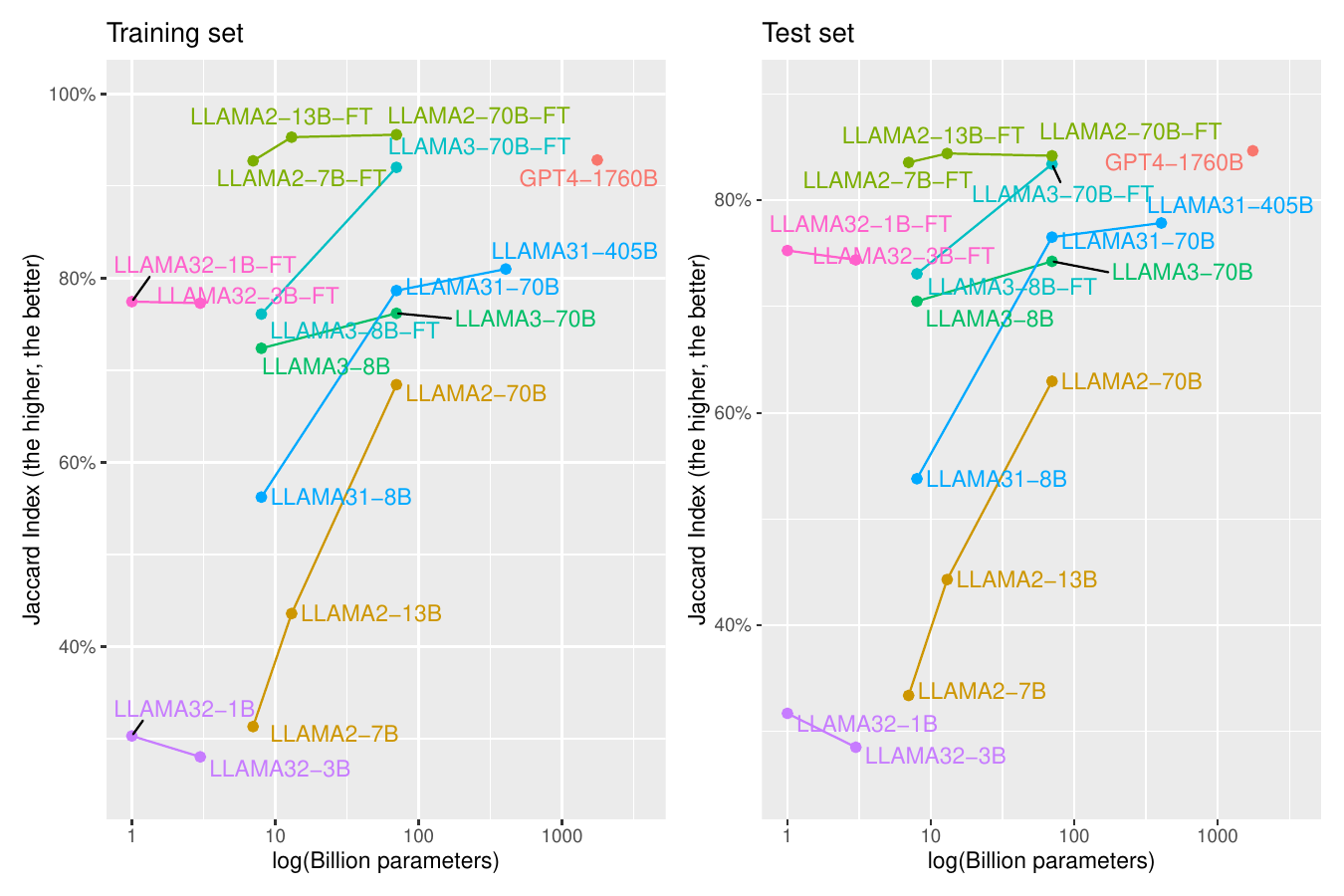}
\caption{Jaccard Index for different LLAMA models and ChatGPT4 on both the training and the test sets. The `FT' suffix stands for `fine-tuned version' of the same model. Number of model parameters in the log-scale. For ChatGPT the real dimension of the model is not known, but seems to be built on top of 8 models of about 220B parameters.}
\label{fig:performance3}
\end{figure}

As explained in the above, we generated curated labeled data using the workflow in Figure~\ref{fig:workflow}. We split the data into a training and test set of 2,404 and 2,177 tweets respectively.
Considering that each tweet can be classified along 46 dimensions simultaneously, the total number of codings to analyse are 110,584 and 100,142 respectively for the two sets, meaning that each tweet is replicated according to the number of the flourishing dimensions that apply.
In this application it is important to correctly identify the dimensions that applies but also those that does not apply, i.e. both true positives and true negatives matter in the analysis. This is why for this analysis we will mainly rely on the Accuracy (in the sense of Jaccard Index) and the Hamming Loss. Figure~\ref{fig:performance1} summarizes the results for the Accuracy, Figure~\ref{fig:performance2} reports the Hamming Loss and Figure~\ref{fig:performance3} the Jaccard Index. 

Before going into the details, we can summarize the evidence as follows:
\begin{itemize}
    \item the larger the number of parameters, the better the performance of the model;
    \item given the size of the model parameters, fine-tuning produces a significant increase in the performances;
    \item give the size of the model, fine-tuning is less effective if the base model has been trained on larger data. 
\end{itemize}
The last fact is relevant for our application. For example, while LLAMA-2 has been trained on 1.5T tokens, LLAMA-3 has been trained on 15T tokens and LLAMA-3.2 on 9T tokens. This seems to imply, looking at the empirical results, that fine-tuning has a hard time in re-weighting  model parameters for those models which are more robust like LLAMA-3 compared to weaker models like LLAMA-2, but for this reason LLAMA-2 7B can be easily fine-tuned and obtain results as good or better than ChatGPT4 (a way much larger parameters model).

Another fact that we notice is that, as expected, the performance of the fine-tuned models are better on the training data than on the test data although the performance of the test data is quite remarkable for such a challenging problem.

All the above results are consistent across Accuracy, Hamming Loss and Jaccard Index.

\subsection*{Attribution of European Parliamentary Questions to Policy Dimension}
As it can be seen in Table~\ref{tab:CAP_PE}, after human validation (Step 3 of workflow in Figure~\ref{fig:workflow}) the zero shot method for ChatGPT4 gives very low Accuracy: 36.6\%.
The base LLAMA2 7B, also with zero shot method, works as badly as ChatGPT4.
The proposed iterative method for ChatGPT4 gets closer to human performance: 75.8\%. This is considered a good result for the CAP project.
The direct method performance has not been reported because we couldn't get any result out of it.

\begin{figure}[ht]
\centering
\includegraphics[width=\linewidth]{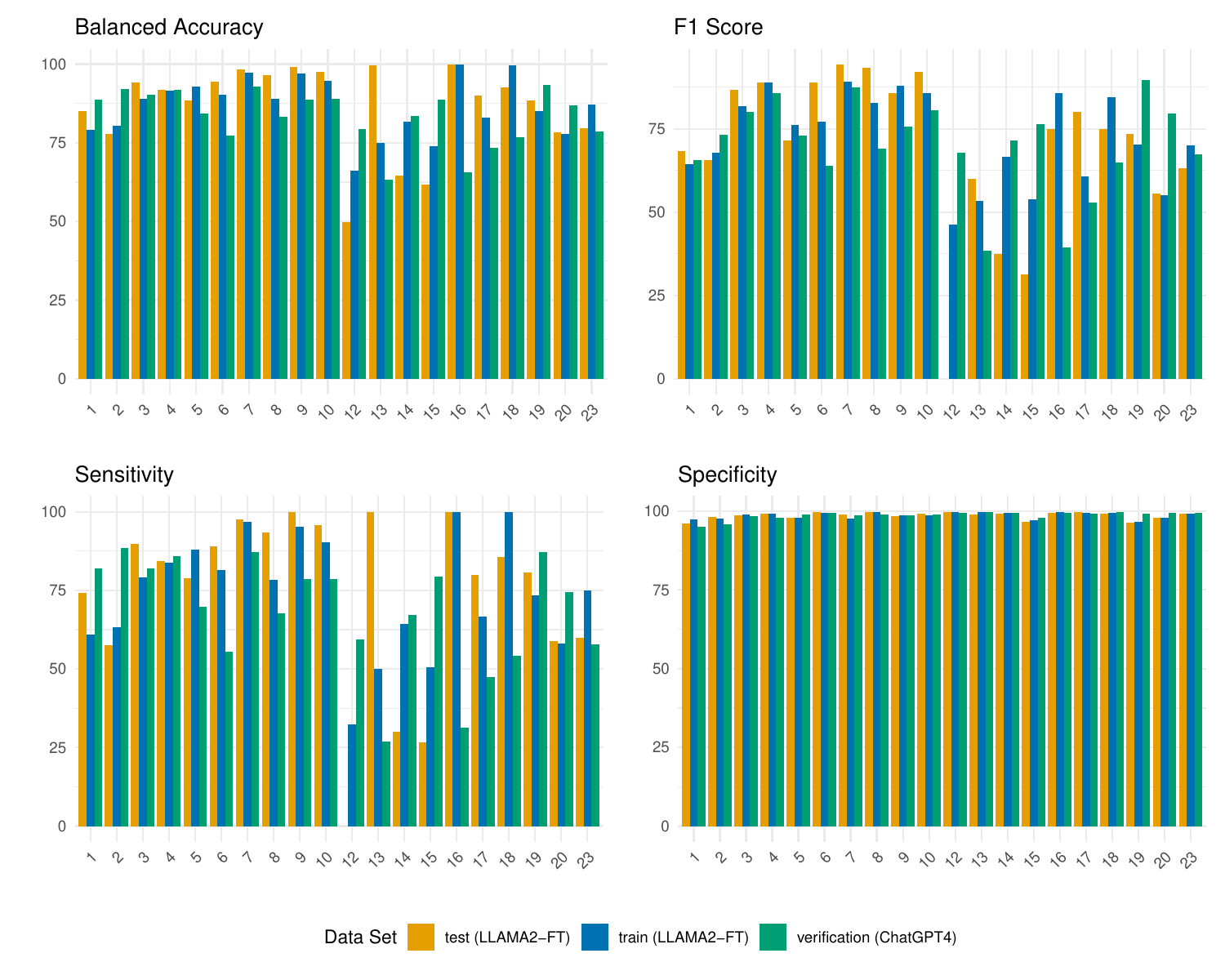}
\caption{Balanced accuracy, F1 score, Sensitivity and Specificity by major policy areas of the Comparative Agenda Project, for the classification of the both the train and test set using the LLAMA-2 fine-tuned model and on the verified set classified by ChatGPT4.}
\label{fig:cap_stats}
\end{figure}

Most surprisingly, after fine-tuning the LLAMA2-7B model (Step 4 of the workflow in Figure~\ref{fig:workflow}), the Accuracy of the zero-shot method jumps from about 37\% to about 75\%.
Figure~\ref{fig:cap_stats} also reports the F1 score, Sensitivity, Specificity and the Balanced Accuracy for the LLAMA2-7B fine-tuned model on both the train and test sets.
Figure~\ref{fig:cap_stats} shows that the Sensitivity is low (less than 50\%) for the policy areas: 12 = `Law and Crime', 14 = `Regional and Urban issues and Planning', and 15 = `Banking, finance and domestic commerce issues'. Policy area 15 is often misclassified as `Macroeconomics', while for the other two areas there is no clear pattern of misclassification.  Further training on those categories might help the accuracy of the fine-tuned model but this is the object of future research although this seems in line with the results found \parencite{SebHok2024} obtained with XMLRoberta (much smaller LLM) trained on a a much larger training set (about 1.6M texts). ChatGPT4 also has a similar behaviour in other classes.
Nevertheless, the Macro Balance Accuracy for the train and test set are, respectively, equal to 86.5\% and 86.4\% for LLAMA2-7B and 83.4\% for ChatGPT4 as shown in Table~\ref{tab:CAP_PE_stats}. 
\begin{table}[ht]
\centering
\begin{tabular}{|c|c|c|c|c|c|c|c|}
\hline
Model & N.Par. & Verification set & Acc. &Training set size & Acc. & Test set size & Acc. \\
\hline
ChatGPT4 (zero shot) & 1,760B &1,964 & 36.6& -- & --  &  & --\\
ChatGPT4 (iterative) & 1,760B &1,964 & 75.8& -- & --  &  & --\\
LLAMA2 & 7B & -- & -- & 1,272 & 37.1& 424 & 37.0 \\
LLAMA2 fine-tuned &7B & -- & -- &1,272  & 75.4 &424  & 74.8  \\
LLAMA2 & 7B & -- & -- & 1,962 & 37.4 & 654  &  37.8 \\
LLAMA2 fine-tuned &7B & -- & -- & 1,962  & 81.1 & 654  & 74.2  \\
\hline
\end{tabular}
\caption{\label{tab:CAP_PE}Accuracy of classification of the CAP project data for different models.}
\end{table}

\begin{table}[ht]
\centering
\begin{tabular}{|c|c|c|c|c|c|c|c|}
\hline
Model &  Set & Size &Accuracy & F1 & Balanced Accuracy & Sensitivity & Specificity \\
\hline
ChatGPT4 (iterative) & verification & 1,964&75.8 &70.0 & 83.4 & 68.1 & 98.7\\
LLAMA2 fine-tuned & train &1,272 & 75.4&72.4 & 86.5 & 74.4 & 98.7 \\
LLAMA2 fine-tuned & test &424& 74.8 &73.0 & 86.4 &  74.1 & 98.7\\
LLAMA2 fine-tuned & train &1,962& 84.2 &81.1 & 91.1 & 83.1 & 99.2 \\
LLAMA2 fine-tuned & test &654 & 74.2 &71.8 & 86.5 &  74.4 & 98.6\\
\hline
\end{tabular}
\caption{\label{tab:CAP_PE_stats}Overall Accuracy and average values of F1 score, Balanced Accuracy, Sensitivity and Specificity in the classification of the CAP project data for different models and data sets.}
\end{table}

\subsection*{Harvard Dataverse Dataset}
Table~\ref{tab:DV1} reports the  accuracy of the LLAMA2 family of models in this multi-label classification task. As it can be seen, a fine-tuned 7B model can do as good as a 70B parameter model with as little as 5,000 labeled sets, i.e. both reach an accuracy of about 84\%. By extending the training set to its whole size of 76,110 records, the accuracy get as high as 94.6\%.
\begin{table}[ht]
\centering
\begin{tabular}{|c|c|c|c|c|}
\hline
LLAMA2 model & Parameters & Training set size & Test set size &  \% at least one correct \\
\hline
base & 7B & -- & 32,619  & 54.4 \\
base &13B & -- &32,619  & 54.3 \\
base &70B & -- &32,619  & 84.9 \\
fine-tuned (small) &7B & 5,000 & 32,619  & 84.8 \\
fine-tuned (large) &7B & 76,110 & 32,619  & 94.6 \\
\hline
\end{tabular}
\caption{\label{tab:DV1}Accuracy of at least one subject category predicted correctly as a function of number of model parameters, fine-tuning and size of training set.}
\end{table}
It is interesting to notice that, even though the LLM is asked to produce at most three subject categories, in a few cases it proposes more labels than those considered \textit{true}. Although we could not run an extensive human supervised checking, a quick inspection shows that in many cases these additional categories seem to apply, especially in the case when the dataset was originally labeled according to only one subject category. Further investigation is in the plans, because this could be the real added value of using generative AI in this application.
Table~\ref{tab:DV2} shows the accuracy in predicting the correct number of categories: 90.3\% of the times the fine-tuned (large) model predicts exactly the same categories as those existing, and almost symmetrically it misses or produce more entries.
The table cuts out the rows for which the number of true categories is 5,6, 8 and 14 (a total of 34 datasets, 0.1\% of the data) and the columns for which the predict number of categories is 5, 6, or 7 (27 datasets in total or 0.08\% of the datasets).
\begin{table}[ht]
\centering
\begin{tabular}{|cl|r|r|r|r|}
\hline
&Predicted   & 1 & 2 & 3 & 4 \\
True   &&  &  &  & \\
\hline
1 && 80.3 & 3.4 & 0.7 & 0.1 \\
2 && 3.6 & 9.8 & 0.3 & 0.0 \\
3 && 0.6 & 0.4 & 0.2 & 0.0 \\
4 && 0.1 & 0.1 & 0.0 & 0.0\\
\hline
\end{tabular}
\caption{\label{tab:DV2}Accuracy of (large) fine-tuned model in predicting the exact categories (and number of categories). The table cuts out the rows for which the number of true categories is 5,6, 8 and 14 (a total of 34 datasets, 0.1\% of the data) and the columns for which the predict number of categories is 5, 6, or 7 (27 datasets in total or 0.08\% of the datasets). Overall, the Accuracy in predicting exactly 1, 2 and 3 categories is 90.3\%.}
\end{table}

\section*{Discussion}

In this study, we demonstrate that small, fine-tuned open-source language models, such as LLAMA2-7B, can achieve performance comparable or better than large commercial models like ChatGPT-4. Our findings underscore that, while very large, closed-source LLMs often excel in text classification tasks, their use comes with substantial drawbacks—particularly for large-scale social science applications. Closed-source models pose risks related to transparency, data security, and replicability and rely on proprietary systems, which can impede open scientific inquiry and drive up costs. In contrast, open-source models, although requiring more careful setup, offer distinct advantages by allowing local deployment for data privacy, task-specific fine-tuning, ease of sharing, and seamless integration into reproducible workflows.

We also highlight the efficacy of a hybrid approach to dataset creation that leverages the strengths of both LLMs and human oversight, thus bypassing some of the traditional limitations of manual text classification. The conventional process—relying on human coders to label documents—often involves high rates of coder disagreement (up to 15-20\%) due to subjective biases, rigid codebooks, or coder fatigue. This method not only introduces variability but also slows down the labeling process significantly, which can become prohibitively expensive in high-dimensional classification tasks with numerous categories. Emerging research further questions the reliability of human annotations, which have traditionally been considered the ``gold standard'' for text classification.

Our proposed workflow circumvents these issues by using multiple LLMs to provide initial classifications and then relying on human reviewers to reject only the labels that do not apply, rather than selecting the correct ones. This approach leverages cognitive strengths—such as error salience and negativity bias—that make it easier for humans to detect errors rather than assign precise labels. As a result, this hybrid process is faster, more efficient, and reduces the demand on human coders, particularly in high-dimensional classification contexts where coders might otherwise face significant cognitive load.

This workflow comprises five key steps: (1) data normalization for a consistent tabular format, (2) AI-driven ``crowd'' classification using one or more LLMs to produce initial labels for each document, (3) human verification to filter out incorrect labels, (4) fine-tuning of a single LLM on the human-filtered labeled dataset, and (5) scaling to classify additional, previously unseen documents. This streamlined approach to labeled dataset creation makes it possible to fine-tune smaller models efficiently and expand the classification process to large datasets, ensuring high-quality, reproducible results without compromising data sensitivity.

Furthermore, our results reveal that smaller, fine-tuned models like LLAMA2-7B not only reach but often match or surpass the performance of large models like ChatGPT-4 on specific tasks. While large models are robust across general applications due to extensive training on broad datasets, this generalist nature can make them less adaptable to domain-specific tasks, as their weights are less responsive to fine-tuning adjustments. In contrast, smaller models tailored with a focused training set can adapt more responsively to specialized tasks, balancing the need for precision and scalability in applied research contexts.

In conclusion, this study demonstrates that, given a robust training set and a streamlined hybrid workflow, smaller, fine-tuned open-source models are not only viable but advantageous for specific applications in social science research. Such models provide comparable performance to larger, closed models while offering significant benefits in transparency, adaptability, data sensitivity, and computational efficiency. This approach underscores the potential for smaller, open-source models to support high-performance, reproducible research, with a cost-effective and privacy-conscious framework that empowers researchers to conduct scalable, collaborative studies.


	\printbibliography

@article{Harsay2012, 
    author = {Harsay, H.A. and Spaan, M. and Wijnen, J.G. and Ridderinkhof, K.R.},
    title = {Error awareness and salience processing in the oddball task: shared neural mechanisms},
    journal = {Front. Hum. Neurosci.},
    year = {2012},
    volume = {6},
    issue = {246}
}

@article{lazer2009,
author = {David Lazer  and Alex Pentland  and Lada Adamic  and Sinan Aral  and Albert-László Barabási  and Devon Brewer  and Nicholas Christakis  and Noshir Contractor  and James Fowler  and Myron Gutmann  and Tony Jebara  and Gary King  and Michael Macy  and Deb Roy  and Marshall Van Alstyne },
title = {Computational Social Science},
journal = {Science},
volume = {323},
number = {5915},
pages = {721-723},
year = {2009},
doi = {10.1126/science.1167742},
URL = {https://www.science.org/doi/abs/10.1126/science.1167742},
eprint = {https://www.science.org/doi/pdf/10.1126/science.1167742}}

@article{Hagendorff2023,
	author = {Hagendorff, Thilo and Fabi, Sarah and Kosinski, Michal},
	date = {2023/10/01},
	doi = {10.1038/s43588-023-00527-x},
	id = {Hagendorff2023},
	isbn = {2662-8457},
	journal = {Nature Computational Science},
	number = {10},
	pages = {833--838},
	title = {Human-like intuitive behavior and reasoning biases emerged in large language models but disappeared in ChatGPT},
	url = {https://doi.org/10.1038/s43588-023-00527-x},
	volume = {3},
	year = {2023}}

@article{Raffel2020,
author = {Raffel, Colin and Shazeer, Noam and Roberts, Adam and Lee, Katherine and Narang, Sharan and Matena, Michael and Zhou, Yanqi and Li, Wei and Liu, Peter J.},
title = {Exploring the limits of transfer learning with a unified text-to-text transformer},
year = {2020},
issue_date = {January 2020},
publisher = {JMLR.org},
volume = {21},
number = {1},
issn = {1532-4435},
journal = {J. Mach. Learn. Res.},
month = jan,
articleno = {140},
numpages = {67}
}

@inproceedings{brown2020,
 author = {Brown, Tom and Mann, Benjamin and Ryder, Nick and Subbiah, Melanie and Kaplan, Jared D and Dhariwal, Prafulla and Neelakantan, Arvind and Shyam, Pranav and Sastry, Girish and Askell, Amanda and Agarwal, Sandhini and Herbert-Voss, Ariel and Krueger, Gretchen and Henighan, Tom and Child, Rewon and Ramesh, Aditya and Ziegler, Daniel and Wu, Jeffrey and Winter, Clemens and Hesse, Chris and Chen, Mark and Sigler, Eric and Litwin, Mateusz and Gray, Scott and Chess, Benjamin and Clark, Jack and Berner, Christopher and McCandlish, Sam and Radford, Alec and Sutskever, Ilya and Amodei, Dario},
 booktitle = {Advances in Neural Information Processing Systems},
 editor = {H. Larochelle and M. Ranzato and R. Hadsell and M.F. Balcan and H. Lin},
 pages = {1877--1901},
 publisher = {Curran Associates, Inc.},
 title = {Language Models are Few-Shot Learners},
 url = {https://proceedings.neurips.cc/paper_files/paper/2020/file/1457c0d6bfcb4967418bfb8ac142f64a-Paper.pdf},
 volume = {33},
 year = {2020}
}

@article{Gilardi2023,
author = {Fabrizio Gilardi  and Meysam Alizadeh  and Maël Kubli },
title = {ChatGPT outperforms crowd workers for text-annotation tasks},
journal = {Proceedings of the National Academy of Sciences},
volume = {120},
number = {30},
pages = {e2305016120},
year = {2023},
doi = {10.1073/pnas.2305016120},
URL = {https://www.pnas.org/doi/abs/10.1073/pnas.2305016120},
eprint = {https://www.pnas.org/doi/pdf/10.1073/pnas.2305016120}}

@article{ito1998negative,
  title={Negative information weighs more heavily on the brain: The negativity bias in evaluative categorizations},
  author={Ito, Tiffany A. and Larsen, Jeff T. and Smith, Nicole K. and Cacioppo, John T.},
  journal={Journal of Personality and Social Psychology},
  volume={75},
  number={4},
  pages={887--900},
  year={1998},
  publisher={American Psychological Association},
  doi={10.1037//0022-3514.75.4.887}
}

@article{Mobbs2015,
title = {The ecology of human fear: survival optimization and the nervous system},
author = {Mobbs, Paul D. and Hagan, Christopher C.},
journal = {Frontiers in Neuroscience},
volume = {9},
pages = {55},
year = {2015},
doi = {10.3389/fnins.2015.00055}
}

@article{Singh2024,
	author = {Singh, Gopendra Vikram and Ghosh, Soumitra and Firdaus, Mauajama and Ekbal, Asif and Bhattacharyya, Pushpak},
	date = {2024/05/28},
	doi = {10.1038/s41598-024-58944-5},
	id = {Singh2024},
	isbn = {2045-2322},
	journal = {Scientific Reports},
	number = {1},
	pages = {12204},
	title = {Predicting multi-label emojis, emotions, and sentiments in code-mixed texts using an emojifying sentiments framework},
	url = {https://doi.org/10.1038/s41598-024-58944-5},
	volume = {14},
	year = {2024}}

@inproceedings{
hanke2024open,
title={Open {LLM}s are Necessary for Private Adaptations and Outperform their Closed Alternatives},
author={Vincent Hanke and Tom Blanchard and Franziska Boenisch and Iyiola Emmanuel Olatunji and Michael Backes and Adam Dziedzic},
booktitle={ICML 2024 Next Generation of AI Safety Workshop},
year={2024},
url={https://openreview.net/forum?id=uGml3wUL8s}
}

@inproceedings{Vaswani17,
author = {Vaswani, Ashish and Shazeer, Noam and Parmar, Niki and Uszkoreit, Jakob and Jones, Llion and Gomez, Aidan N. and Kaiser, \L{}ukasz and Polosukhin, Illia},
title = {Attention is all you need},
year = {2017},
isbn = {9781510860964},
publisher = {Curran Associates Inc.},
address = {Red Hook, NY, USA},
booktitle = {Proceedings of the 31st International Conference on Neural Information Processing Systems},
pages = {6000–6010},
numpages = {11},
location = {Long Beach, California, USA},
series = {NIPS'17}
}

@inproceedings{conneau-etal-2020-unsupervised,
    title = "Unsupervised Cross-lingual Representation Learning at Scale",
    author = "Conneau, Alexis  and
      Khandelwal, Kartikay  and
      Goyal, Naman  and
      Chaudhary, Vishrav  and
      Wenzek, Guillaume  and
      Guzm{\'a}n, Francisco  and
      Grave, Edouard  and
      Ott, Myle  and
      Zettlemoyer, Luke  and
      Stoyanov, Veselin",
    editor = "Jurafsky, Dan  and
      Chai, Joyce  and
      Schluter, Natalie  and
      Tetreault, Joel",
    booktitle = "Proceedings of the 58th Annual Meeting of the Association for Computational Linguistics",
    month = jul,
    year = "2020",
    address = "Online",
    publisher = "Association for Computational Linguistics",
    url = "https://aclanthology.org/2020.acl-main.747",
    doi = "10.18653/v1/2020.acl-main.747",
    pages = "8440--8451",
}

@inproceedings{devlin-etal-2019-bert,
    title = "{BERT}: Pre-training of Deep Bidirectional Transformers for Language Understanding",
    author = "Devlin, Jacob  and
      Chang, Ming-Wei  and
      Lee, Kenton  and
      Toutanova, Kristina",
    editor = "Burstein, Jill  and
      Doran, Christy  and
      Solorio, Thamar",
    booktitle = "Proceedings of the 2019 Conference of the North {A}merican Chapter of the Association for Computational Linguistics: Human Language Technologies, Volume 1 (Long and Short Papers)",
    month = jun,
    year = "2019",
    address = "Minneapolis, Minnesota",
    publisher = "Association for Computational Linguistics",
    url = "https://aclanthology.org/N19-1423",
    doi = "10.18653/v1/N19-1423",
    pages = "4171--4186",
}

@article{Suhaib24,
    author = {Abdurahman, Suhaib and Atari, Mohammad and Karimi-Malekabadi, Farzan and Xue, Mona J and Trager, Jackson and Park, Peter S and Golazizian, Preni and Omrani, Ali and Dehghani, Morteza},
    title = "{Perils and opportunities in using large language models in psychological research}",
    journal = {PNAS Nexus},
    volume = {3},
    number = {7},
    pages = {pgae245},
    year = {2024},
    month = {07},
    issn = {2752-6542},
    doi = {10.1093/pnasnexus/pgae245},
    url = {https://doi.org/10.1093/pnasnexus/pgae245},
    eprint = {https://academic.oup.com/pnasnexus/article-pdf/3/7/pgae245/58645406/pgae245.pdf},
}

@article{Nowak24-german,
	author = {Nowak, Sebastian and Sprinkart, Alois M.},
	date = {2024/10/01},
	doi = {10.1007/s00117-024-01327-8},
	id = {Nowak2024},
	isbn = {2731-7056},
	journal = {Die Radiologie},
	number = {10},
	pages = {779--786},
	title = {Gro{\ss}e Sprachmodelle von OpenAI, Google, Meta, X und Co.},
	url = {https://doi.org/10.1007/s00117-024-01327-8},
	volume = {64},
	year = {2024}}

@ARTICLE{Nowak2024,
  title     = "Large language models from {OpenAI}, Google, Meta, {X} and Co. :
               The role of ``closed'' and ``open'' models in radiology",
  author    = "Nowak, Sebastian and Sprinkart, Alois M",
  journal   = "Radiologie (Heidelb.)",
  publisher = "Springer Science and Business Media LLC",
  volume    =  64,
  number    =  10,
  pages     = "779--786",
  month     =  oct,
  year      =  2024,
  copyright = "https://www.springernature.com/gp/researchers/text-and-data-mining",
  language  = "de"
}

@article{Miah2024,
	author = {Miah, Md Saef Ullah and Kabir, Md Mohsin and Sarwar, Talha Bin and Safran, Mejdl and Alfarhood, Sultan and Mridha, M. F.},
	date = {2024/04/26},
	doi = {10.1038/s41598-024-60210-7},
	id = {Miah2024},
	isbn = {2045-2322},
	journal = {Scientific Reports},
	number = {1},
	pages = {9603},
	title = {A multimodal approach to cross-lingual sentiment analysis with ensemble of transformer and LLM},
	url = {https://doi.org/10.1038/s41598-024-60210-7},
	volume = {14},
	year = {2024}}

@article{Zhang24,
	author = {Zhang, Gongbo and Jin, Qiao and Zhou, Yiliang and Wang, Song and Idnay, Betina and Luo, Yiming and Park, Elizabeth and Nestor, Jordan G. and Spotnitz, Matthew E. and Soroush, Ali and Campion, Thomas R. and Lu, Zhiyong and Weng, Chunhua and Peng, Yifan},
	date = {2024/09/09},
	doi = {10.1038/s41746-024-01239-w},
	id = {Zhang2024},
	isbn = {2398-6352},
	journal = {npj Digital Medicine},
	number = {1},
	pages = {239},
	title = {Closing the gap between open source and commercial large language models for medical evidence summarization},
	url = {https://doi.org/10.1038/s41746-024-01239-w},
	volume = {7},
	year = {2024}}

@misc{Lewis16,
author = {Lewis, Benjamin and Kakkar, Devika},
publisher = {Harvard Dataverse},
title = {{Harvard CGA Geotweet Archive v2.0}},
year = {2016},
version = {V2},
doi = {10.7910/DVN/3NCMB6},
url = {https://doi.org/10.7910/DVN/3NCMB6}
}

@article{Alexandrova14,
author = {Petya Alexandrova and Marcello Carammia and Sebastian Princen and Arco Timmermans},
title ={Measuring the European Council agenda: Introducing a new approach and dataset},

journal = {European Union Politics},
volume = {15},
number = {1},
pages = {152-167},
year = {2014},
doi = {10.1177/1465116513509124},

URL = { 
    
        https://doi.org/10.1177/1465116513509124
    
    

},
eprint = { 
    
        https://doi.org/10.1177/1465116513509124
    
    

}
}

@article{Gougherty2024,
	author = {Gougherty, Andrew V. and Clipp, Hannah L.},
	date = {2024/05/16},
	doi = {10.1038/s44185-024-00043-9},
	id = {Gougherty2024},
	isbn = {2731-4243},
	journal = {npj Biodiversity},
	number = {1},
	pages = {13},
	title = {Testing the reliability of an AI-based large language model to extract ecological information from the scientific literature},
	url = {https://doi.org/10.1038/s44185-024-00043-9},
	volume = {3},
	year = {2024}}

@misc{TurboCurator,
  author       = {ICPSR},
  title        = {TurboCurator: Enhancing Metadata Quality for Data Depositors},
  year         = {2023},
  url          = {https://turbocurator.icpsr.umich.edu/tc/about},
  note         = {Accessed: 2024-10-15},
  institution  = {Inter-university Consortium for Political and Social Research (ICPSR), University of Michigan}
}

@article{sdata201618,
	author = {Wilkinson, Mark D. and Dumontier, Michel and Aalbersberg, IJsbrand Jan and Appleton, Gabrielle and Axton, Myles and Baak, Arie and Blomberg, Niklas and Boiten, Jan-Willem and da Silva Santos, Luiz Bonino and Bourne, Philip E. and Bouwman, Jildau and Brookes, Anthony J. and Clark, Tim and Crosas, Merc{\`e} and Dillo, Ingrid and Dumon, Olivier and Edmunds, Scott and Evelo, Chris T. and Finkers, Richard and Gonzalez-Beltran, Alejandra and Gray, Alasdair J. G. and Groth, Paul and Goble, Carole and Grethe, Jeffrey S. and Heringa, Jaap and 't Hoen, Peter A. C and Hooft, Rob and Kuhn, Tobias and Kok, Ruben and Kok, Joost and Lusher, Scott J. and Martone, Maryann E. and Mons, Albert and Packer, Abel L. and Persson, Bengt and Rocca-Serra, Philippe and Roos, Marco and van Schaik, Rene and Sansone, Susanna-Assunta and Schultes, Erik and Sengstag, Thierry and Slater, Ted and Strawn, George and Swertz, Morris A. and Thompson, Mark and van der Lei, Johan and van Mulligen, Erik and Velterop, Jan and Waagmeester, Andra and Wittenburg, Peter and Wolstencroft, Katherine and Zhao, Jun and Mons, Barend},
	date = {2016/03/15},
	doi = {10.1038/sdata.2016.18},
	id = {Wilkinson2016},
	isbn = {2052-4463},
	journal = {Scientific Data},
	number = {1},
	pages = {160018},
	title = {The \uppercase{FAIR} Guiding Principles for scientific data management and stewardship},
	url = {https://doi.org/10.1038/sdata.2016.18},
	volume = {3},
	year = {2016}}

@misc{shigapov2024,
      title={FAIR GPT: A virtual consultant for research data management in ChatGPT}, 
      author={Renat Shigapov and Irene Schumm},
      year={2024},
      eprint={2410.07108},
      archivePrefix={arXiv},
      primaryClass={cs.DL},
      url={https://arxiv.org/abs/2410.07108}, 
}

@article {King07,
	title = {An Introduction to the Dataverse Network as an Infrastructure for Data Sharing},
	journal = {Sociological Methods and Research},
	volume = {36},
	year = {2007},
	pages = {173{\textendash}199},
	author = {Gary King}
}

@article{VanderWeele2017,
author = {Tyler J. VanderWeele },
title = {On the promotion of human flourishing},
journal = {Proceedings of the National Academy of Sciences},
volume = {114},
number = {31},
pages = {8148-8156},
year = {2017},
doi = {10.1073/pnas.1702996114},
URL = {https://www.pnas.org/doi/abs/10.1073/pnas.1702996114},
eprint = {https://www.pnas.org/doi/pdf/10.1073/pnas.1702996114}}

@misc{gpt4all,
  author = {Yuvanesh Anand and Zach Nussbaum and Brandon Duderstadt and Benjamin Schmidt and Andriy Mulyar},
  title = {GPT4All: Training an Assistant-style Chatbot with Large Scale Data Distillation from GPT-3.5-Turbo},
  year = {2023},
  publisher = {GitHub},
  journal = {GitHub repository},
  howpublished = {\url{https://github.com/nomic-ai/gpt4all}},
}

@misc{ollama,
  author = {Michael Chiang},
  title = {Get up and running with Llama 3.2, Mistral, Gemma 2, and other large language models},
  year = {2023},
  publisher = {GitHub},
  journal = {GitHub repository},
  howpublished = {\url{https://github.com/ollama/ollama}},
}

@article{Gilbert1884,
   author    =  "G. K. Gilbert",
   title     =  "Finley’s tornado predictions",
   year      =  "1884",
   journal   =  "Amer. Meteor. J.",
   volume    =  "1",
   pages     =  "166--172"
}

@Article{Jaccard1901,
author={Jaccard, Paul},
title={{\'E}tude comparative de la distribution florale dans une portion des Alpes et du Jura},
journal={Bulletin de la Soci{\'e}t{\'e} Vaudoise des Sciences Naturelles},
year={1901},
publisher={Imprimerie Corbaz {\&} Comp.},
volume={37},
number={142},
pages={547},
issn={0037-9603},
doi={10.5169/seals-266450},
url={https://doi.org/10.5169/seals-266450}
}

@article{Hamming50,
  author={Hamming, R. W.},
  journal={The Bell System Technical Journal}, 
  title={Error detecting and error correcting codes}, 
  year={1950},
  volume={29},
  number={2},
  pages={147-160},
  doi={10.1002/j.1538-7305.1950.tb00463.x}}

@misc{llamacpp,
  author = {Georgi Gerganov},
  title = {LLM inference in C/C++},
  year = {2022},
  publisher = {GitHub},
  journal = {GitHub repository},
  howpublished = {\url{https://github.com/ggerganov/llama.cpp}},
}

@misc{CAP,
      title={Policy Agendas Project: Codebook}, 
      author={Jones, Bryan D. and Frank R. Baumgartner and Sean M. Theriault and Derek A. Epp and Cheyenne Lee and Miranda E. Sullivan},
      year={2023},
      howpublished = {\url{https://www.comparativeagendas.net}}
}

@misc{hu2021,
      title={LoRA: Low-Rank Adaptation of Large Language Models}, 
      author={Edward J. Hu and Yelong Shen and Phillip Wallis and Zeyuan Allen-Zhu and Yuanzhi Li and Shean Wang and Lu Wang and Weizhu Chen},
      year={2021},
      eprint={2106.09685},
      archivePrefix={arXiv},
      primaryClass={cs.CL},
      url={https://arxiv.org/abs/2106.09685}, 
}

@INPROCEEDINGS{Wei23,
  author={Wei, Fusheng and Keeling, Robert and Huber-Fliflet, Nathaniel and Zhang, Jianping and Dabrowski, Adam and Yang, Jingchao and Mao, Qiang and Qin, Han},
  booktitle={2023 IEEE International Conference on Big Data (BigData)}, 
  title={Empirical Study of LLM Fine-Tuning for Text Classification in Legal Document Review}, 
  year={2023},
  volume={},
  number={},
  pages={2786-2792},
  doi={10.1109/BigData59044.2023.10386911}}

@INPROCEEDINGS{Nikolakopoulos24,
  author={Nikolakopoulos, Anastasios and Evangelatos, Spyridon and Veroni, Eleni and Chasapas, Konstantinos and Gousetis, Nikolaos and Apostolaras, Apostolos and Nikolopoulos, Christos D. and Korakis, Thanasis},
  booktitle={2024 5th International Conference in Electronic Engineering, Information Technology \& Education (EEITE)}, 
  title={Large Language Models in Modern Forensic Investigations: Harnessing the Power of Generative Artificial Intelligence in Crime Resolution and Suspect Identification}, 
  year={2024},
  volume={},
  number={},
  pages={1-5},
  doi={10.1109/EEITE61750.2024.10654427}}

@misc{kozlowski2024,
      title={In Silico Sociology: Forecasting COVID-19 Polarization with Large Language Models}, 
      author={Austin C. Kozlowski and Hyunku Kwon and James A. Evans},
      year={2024},
      eprint={2407.11190},
      archivePrefix={arXiv},
      primaryClass={cs.CY},
      url={https://arxiv.org/abs/2407.11190}, 
}

@misc{LiuGe2024,
      title={PoliPrompt: A High-Performance Cost-Effective LLM-Based Text Classification Framework for Political Science}, 
      author={Menglin Liu and Ge Shi},
      year={2024},
      eprint={2409.01466},
      archivePrefix={arXiv},
      primaryClass={cs.CL},
      url={https://arxiv.org/abs/2409.01466}, 
}

@article{Guo24,
    author = {Guo, Yuting and Ovadje, Anthony and Al-Garadi, Mohammed Ali and Sarker, Abeed},
    title = "{Evaluating large language models for health-related text classification tasks with public social media data}",
    journal = {Journal of the American Medical Informatics Association},
    volume = {31},
    number = {10},
    pages = {2181-2189},
    year = {2024},
    month = {08},
    issn = {1527-974X},
    doi = {10.1093/jamia/ocae210},
    url = {https://doi.org/10.1093/jamia/ocae210},
    eprint = {https://academic.oup.com/jamia/article-pdf/31/10/2181/59206398/ocae210.pdf},
}

@inproceedings{Loukas23,
author = {Loukas, Lefteris and Stogiannidis, Ilias and Diamantopoulos, Odysseas and Malakasiotis, Prodromos and Vassos, Stavros},
title = {Making LLMs Worth Every Penny: Resource-Limited Text Classification in Banking},
year = {2023},
isbn = {9798400702402},
publisher = {Association for Computing Machinery},
address = {New York, NY, USA},
url = {https://doi.org/10.1145/3604237.3626891},
doi = {10.1145/3604237.3626891},
booktitle = {Proceedings of the Fourth ACM International Conference on AI in Finance},
pages = {392–400},
numpages = {9},
location = {Brooklyn, NY, USA},
series = {ICAIF '23}
}

@ARTICLE{Fields24,
  author={Fields, John and Chovanec, Kevin and Madiraju, Praveen},
  journal={IEEE Access}, 
  title={A Survey of Text Classification With Transformers: How Wide? How Large? How Long? How Accurate? How Expensive? How Safe?}, 
  year={2024},
  volume={12},
  number={},
  pages={6518-6531},
  doi={10.1109/ACCESS.2024.3349952}}

@misc{yin2024,
      title={CrisisSense-LLM: Instruction Fine-Tuned Large Language Model for Multi-label Social Media Text Classification in Disaster Informatics}, 
      author={Kai Yin and Chengkai Liu and Ali Mostafavi and Xia Hu},
      year={2024},
      eprint={2406.15477},
      archivePrefix={arXiv},
      primaryClass={cs.CL},
      url={https://arxiv.org/abs/2406.15477}, 
}

@misc{rouzegar2024,
      title={Enhancing Text Classification through LLM-Driven Active Learning and Human Annotation}, 
      author={Hamidreza Rouzegar and Masoud Makrehchi},
      year={2024},
      eprint={2406.12114},
      archivePrefix={arXiv},
      primaryClass={cs.CL},
      url={https://arxiv.org/abs/2406.12114}, 
}

@misc{pangakis2024,
      title={Knowledge Distillation in Automated Annotation: Supervised Text Classification with LLM-Generated Training Labels}, 
      author={Nicholas Pangakis and Samuel Wolken},
      year={2024},
      eprint={2406.17633},
      archivePrefix={arXiv},
      primaryClass={cs.CL},
      url={https://arxiv.org/abs/2406.17633}, 
}

@article{Wang2024,
	author = {Wang, Yu},
	db = {Cambridge Core},
	doi = {DOI: 10.1017/pan.2023.36},
	dp = {Cambridge University Press},
	edition = {2023/11/28},
	isbn = {1047-1987},
	journal = {Political Analysis},
	pages = {1-5},
	publisher = {Cambridge University Press},
	title = {On Finetuning Large Language Models},
	url = {https://www.cambridge.org/core/product/443356A60ACD5B1716B14DD204BDEA1F},
	year = {forthcoming}}

@article{Chae2023,
  author    = {Youngjin Chae and Thomas Davidson},
  title     = {Large Language Models for Text Classification: From Zero-shot Learning to Instruction-tuning},
  year      = {2023},
  journal   = {SocArXiv},
  month     = {August 24},
  doi       = {10.31235/osf.io/sthwk},
  url       = {https://doi.org/10.31235/osf.io/sthwk}
}

@techreport{OECD_FOS_Classification_2007,
  title        = {Revised Field of Science and Technology (FOS) Classification in the Frascati Manual},
  author       = {{Organisation for Economic Co-operation and Development (OECD)}},
  year         = {2007},
  institution  = {Organisation for Economic Co-operation and Development},
  address      = {Paris},
  url          = {https://www.oecd.org/science/inno/38235147.pdf},
  note         = {Accessed: 2024-10-14}
}

@inproceedings{Clark2021,
    title = "All That{'}s {`}Human{'} Is Not Gold: Evaluating Human Evaluation of Generated Text",
    author = "Clark, Elizabeth  and
      August, Tal  and
      Serrano, Sofia  and
      Haduong, Nikita  and
      Gururangan, Suchin  and
      Smith, Noah A.",
    editor = "Zong, Chengqing  and
      Xia, Fei  and
      Li, Wenjie  and
      Navigli, Roberto",
    booktitle = "Proceedings of the 59th Annual Meeting of the Association for Computational Linguistics and the 11th International Joint Conference on Natural Language Processing (Volume 1: Long Papers)",
    month = aug,
    year = "2021",
    address = "Online",
    publisher = "Association for Computational Linguistics",
    url = "https://aclanthology.org/2021.acl-long.565",
    doi = "10.18653/v1/2021.acl-long.565",
    pages = "7282--7296",
}

@misc{fabbri2021,
      title={SummEval: Re-evaluating Summarization Evaluation}, 
      author={Alexander R. Fabbri and Wojciech Kryściński and Bryan McCann and Caiming Xiong and Richard Socher and Dragomir Radev},
      year={2021},
      eprint={2007.12626},
      archivePrefix={arXiv},
      primaryClass={cs.CL},
      url={https://arxiv.org/abs/2007.12626}, 
}

@misc{Goyal2023,
      title={News Summarization and Evaluation in the Era of GPT-3}, 
      author={Tanya Goyal and Junyi Jessy Li and Greg Durrett},
      year={2023},
      eprint={2209.12356},
      archivePrefix={arXiv},
      primaryClass={cs.CL},
      url={https://arxiv.org/abs/2209.12356}, 
}

@inproceedings{Liu23,
    title = "Revisiting the Gold Standard: Grounding Summarization Evaluation with Robust Human Evaluation",
    author = "Liu, Yixin  and
      Fabbri, Alex  and
      Liu, Pengfei  and
      Zhao, Yilun  and
      Nan, Linyong  and
      Han, Ruilin  and
      Han, Simeng  and
      Joty, Shafiq  and
      Wu, Chien-Sheng  and
      Xiong, Caiming  and
      Radev, Dragomir",
    editor = "Rogers, Anna  and
      Boyd-Graber, Jordan  and
      Okazaki, Naoaki",
    booktitle = "Proceedings of the 61st Annual Meeting of the Association for Computational Linguistics (Volume 1: Long Papers)",
    month = jul,
    year = "2023",
    address = "Toronto, Canada",
    publisher = "Association for Computational Linguistics",
    url = "https://aclanthology.org/2023.acl-long.228",
    doi = "10.18653/v1/2023.acl-long.228",
    pages = "4140--4170",
}

@article{Kiyasseh24,
	author = {Kiyasseh, Dani and Cohen, Aaron and Jiang, Chengsheng and Altieri, Nicholas},
	date = {2024/02/28},
	doi = {10.1038/s41467-024-46000-9},
	id = {Kiyasseh2024},
	isbn = {2041-1723},
	journal = {Nature Communications},
	number = {1},
	pages = {1808},
	title = {A framework for evaluating clinical artificial intelligence systems without ground-truth annotations},
	url = {https://doi.org/10.1038/s41467-024-46000-9},
	volume = {15},
	year = {2024}}

@incollection{Mutzel23,
    author = {M\"utzel, Sophie and Ollion, \'Etienne},
    isbn = {9780197653609},
     year = {2023},
    title = "{Machine Learning and the Analysis of Culture}",
    booktitle = "{The Oxford Handbook of the Sociology of Machine Learning}",
    publisher = {Oxford University Press},
    doi = {10.1093/oxfordhb/9780197653609.013.39},
    url = {https://doi.org/10.1093/oxfordhb/9780197653609.013.39},
    eprint = {https://academic.oup.com/book/0/chapter/467876363/chapter-ag-pdf/59016069/book\_55209\_section\_467876363.ag.pdf},
}

@INPROCEEDINGS{fechner2024,
  author={Fechner, Richard and Dorpinghaus, Jens},
  booktitle={Preprints of Position Papers of the 19th Conference on Computer Science and Intelligence Systems (FedCSIS)}, 
  title={No Train, No Pain? Assessing the Ability of LLMs for Text Classification with no Finetuning}, 
  year={2024},
  volume={},
  number={},
  pages={9-16}}

@INPROCEEDINGS{trajanov2023,
  title = {Comparing the performance of ChatGPT and state-of-the-art climate NLP models on climate-related text classification tasks},
booktitle = {Proceedings of the 4th International Conference on Environmental
Design (ICED2023)},
  author = {Trajanov, Dimitar and Lazarev, Gorgi and Chitkushev, Ljubomir, T. and Vodenska, Irena},
  year = {2023},
  URL = {https://open.bu.edu/handle/2144/48956},
  publisher = {OpenBU}
}

@InProceedings{cao2024,
author="Cao, Mengyuan
and Wang, Hang
and Liu, Xiaoming
and Wu, Jiahao
and Zhao, Mengting",
editor="Xu, Hua
and Chen, Qingcai
and Lin, Hongfei
and Wu, Fei
and Liu, Lei
and Tang, Buzhou
and Hao, Tianyong
and Huang, Zhengxing
and Lei, Jianbo
and Li, Zuofeng
and Zong, Hui",
title="LLM Collaboration PLM Improves Critical Information Extraction Tasks in Medical Articles",
booktitle="Health Information Processing. Evaluation Track Papers",
year="2024",
publisher="Springer Nature Singapore",
address="Singapore",
pages="178--185",
isbn="978-981-97-1717-0"
}

@misc{maze2024,
	author = {Maze, Elie and Farahbakhsh, Reza and Barrallon, Pierre-Emmanuel and Jallais, Pierre},
	doi = {10.2118/221083-MS},
	eprint = {https://onepetro.org/SPEATCE/proceedings-pdf/24ATCE/24ATCE/D021S012R002/3604503/spe-221083-ms.pdf},
	month = {09},
	pages = {D021S012R002},
	series = {SPE Annual Technical Conference and Exhibition},
	title = {{Textual Data Augmentation for NER in Geosciences with LLMs}},
	url = {https://doi.org/10.2118/221083-MS},
	volume = {SPE Annual Technical Conference and Exhibition},
	year = {2024}
}

@misc{li2024,
      title={Deep Learning and LLM-based Methods Applied to Stellar Lightcurve Classification}, 
      author={Yu-Yang Li and Yu Bai and Cunshi Wang and Mengwei Qu and Ziteng Lu and Roberto Soria and Jifeng Liu},
      year={2024},
      eprint={2404.10757},
      archivePrefix={arXiv},
      primaryClass={astro-ph.IM},
      url={https://arxiv.org/abs/2404.10757}
}

@misc{zhang_etal_2024,
      title={Pushing The Limit of LLM Capacity for Text Classification}, 
      author={Yazhou Zhang and Mengyao Wang and Chenyu Ren and Qiuchi Li and Prayag Tiwari and Benyou Wang and Jing Qin},
      year={2024},
      eprint={2402.07470},
      archivePrefix={arXiv},
      primaryClass={cs.CL},
      url={https://arxiv.org/abs/2402.07470}, 
}

@misc{liao2024,
      title={From Words to Molecules: A Survey of Large Language Models in Chemistry}, 
      author={Chang Liao and Yemin Yu and Yu Mei and Ying Wei},
      year={2024},
      eprint={2402.01439},
      archivePrefix={arXiv},
      primaryClass={cs.LG},
      url={https://arxiv.org/abs/2402.01439}, 
}

@article{fu2024,
  title={VITA: Towards Open-Source Interactive Omni Multimodal LLM},
  author={Chaoyou Fu and Haojia Lin and Zuwei Long and Yunhang Shen and Meng Zhao and Yifan Zhang and Xiong Wang and Di Yin and Long Ma and Xiawu Zheng and Ran He and Rongrong Ji and Yunsheng Wu and Caifeng Shan and Xing Sun},
  journal={ArXiv},
  year={2024},
  volume={abs/2408.05211},
  url={https://api.semanticscholar.org/CorpusID:271843279}
}

@INPROCEEDINGS{irugal2024,
  author={Irugalbandara, Chandra and Mahendra, Ashish and Daynauth, Roland and Arachchige, Tharuka Kasthuri and Dantanarayana, Jayanaka and Flautner, Krisztian and Tang, Lingjia and Kang, Yiping and Mars, Jason},
  booktitle={2024 IEEE International Symposium on Performance Analysis of Systems and Software (ISPASS)}, 
  title={Scaling Down to Scale Up: A Cost-Benefit Analysis of Replacing OpenAI's LLM with Open Source SLMs in Production}, 
  year={2024},
  volume={},
  number={},
  pages={280-291},
  doi={10.1109/ISPASS61541.2024.00034}}

@article{xiang_li_2023,
  title={FLM-101B: An Open LLM and How to Train It with \$100K Budget},
  author={Xiang Li and Yiqun Yao and Xin Jiang and Xuezhi Fang and Xuying Meng and Siqi Fan and Peng Han and Jing Li and LI DU and Bowen Qin and Zheng Zhang and Aixin Sun and Yequan Wang},
  journal={ArXiv},
  year={2023},
  volume={abs/2309.03852},
  url={https://api.semanticscholar.org/CorpusID:261582615}
}

@article{mathav2024,
  title={Fine Tuning LLM for Enterprise: Practical Guidelines and Recommendations},
  author={J MathavRaj and VM Kushala and Harikrishna Warrier and Yogesh Gupta},
  journal={ArXiv},
  year={2024},
  volume={abs/2404.10779},
  url={https://api.semanticscholar.org/CorpusID:269188062}
}

@article{bucher2024,
  title={Fine-Tuned 'Small' LLMs (Still) Significantly Outperform Zero-Shot Generative AI Models in Text Classification},
  author={Martin Juan Jos'e Bucher and Marco Martini},
  journal={ArXiv},
  year={2024},
  volume={abs/2406.08660},
  url={https://api.semanticscholar.org/CorpusID:270440709}
}

@article{Bisbee2024, title={Synthetic Replacements for Human Survey Data? The Perils of Large Language Models}, volume={32}, DOI={10.1017/pan.2024.5}, number={4}, journal={Political Analysis}, author={Bisbee, James and Clinton, Joshua D. and Dorff, Cassy and Kenkel, Brenton and Larson, Jennifer M.}, year={2024}, pages={401–416}}

@misc{TrustMeBro,
      title={Replication for Language Models.
Problems, Principles, and Best Practice for Political Science}, 
      author={Christopher Barrie,  Alexis Palmer and Arthur Spirling},
      year={2024},
      url={https://lexipalmer13.github.io/research/}, 
}

@misc{Barrie2024,
      title={Prompt Stability Scoring for Text Annotation with Large Language Models}, 
      author={Christopher Barrie and Elli Palaiologou and Petter Törnberg},
      year={2024},
      eprint={2407.02039},
      archivePrefix={arXiv},
      primaryClass={cs.CL},
      url={https://arxiv.org/abs/2407.02039}, 
}

@article{Palmer24,
	author = {Palmer, Alexis and Smith, Noah A. and Spirling, Arthur},
	date = {2024/01/01},
	doi = {10.1038/s43588-023-00585-1},
	id = {Palmer2024},
	isbn = {2662-8457},
	journal = {Nature Computational Science},
	number = {1},
	pages = {2--3},
	title = {Using proprietary language models in academic research requires explicit justification},
	url = {https://doi.org/10.1038/s43588-023-00585-1},
	volume = {4},
	year = {2024}}

@article{SebHok2024,
  title={Leveraging Open Large Language Models for Multilingual Policy Topic Classification: The Babel Machine Approach},
  author={Seb{\H{o}}k, Mikl{\'o}s and M{\'a}t{\'e}, {\'A}kos and Ring, Orsolya and Kov{\'a}cs, Viktor and Lehoczki, Rich{\'a}rd},
  journal={Social Science Computer Review},
  pages={08944393241259434},
  year={2024},
  publisher={SAGE Publications Sage CA: Los Angeles, CA}
}

@article{fleiss1971mns,
  author = {Fleiss, J.L. and others},
  journal = {Psychological Bulletin},
  number = 5,
  pages = {378--382},
  title = {{Measuring nominal scale agreement among many raters}},
  volume = 76,
  year = 1971
}
	
%

%

\section*{Additional information}
The \textbf{Supplementary Material} will be available soon with prompts, scripts for fine-tuning, verification results, datasets, code-books, and fine-tuned models. A link to the repository will be included in the next revision of the manuscript.





\end{document}